\documentclass[preprint,12pt]{elsarticle}

\usepackage{amssymb}

\usepackage[cmex10]{amsmath}
\usepackage{booktabs}

\usepackage{graphicx}
\usepackage{epstopdf}

\usepackage{multicol}
\usepackage{multirow}

\usepackage{url}

\usepackage[colorlinks]{hyperref}

\usepackage{natbib}

\usepackage{color}
\usepackage{array}

\newcolumntype{L}[1]{>{\raggedright\let\newline\\\arraybackslash\hspace{0pt}}m{#1}}
\newcolumntype{R}[1]{>{\raggedleft\let\newline\\\arraybackslash\hspace{0pt}}m{#1}}
\newcolumntype{C}[1]{>{\centering\let\newline\\\arraybackslash\hspace{0pt}}m{#1}}

\newcolumntype{x}{>\small c}

\journal{Pattern Recognition}

\begin{document}

\begin{frontmatter}



\title{Deep Patch Learning for Weakly Supervised Object Classification and Discovery}


\author[1]{Peng~Tang}
\ead{pengtang@hust.edu.cn}
\author[1]{Xinggang~Wang\corref{cor1}}
\ead{xgwang@hust.edu.cn}
\author[1]{Zilong~Huang}
\ead{hzl@hust.edu.cn}
\author[1]{Xiang~Bai}
\ead{xbai@hust.edu.cn}
\author[1]{Wenyu~Liu}
\ead{liuwy@hust.edu.cn}

\address[1]{School of Electronic Information and Communications, Huazhong University of Science and Technology, Wuhan 430074, China}

\cortext[cor1]{Corresponding author.}

\begin{abstract}
Patch-level image representation is very important for object classification and detection, since it is robust to spatial transformation, scale variation, and cluttered background.
Many existing methods usually require fine-grained supervisions (e.g., bounding-box annotations) to learn patch features, which requires a great effort to label images may limit their potential applications.
In this paper, we propose to learn patch features via weak supervisions, i.e., only image-level supervisions.
To achieve this goal, we treat images as bags and patches as instances to integrate the weakly supervised multiple instance learning constraints into deep neural networks.
Also, our method integrates the traditional multiple stages of weakly supervised object classification and discovery into a unified deep convolutional neural network and optimizes the network in an end-to-end way.
The network processes the two tasks object classification and discovery jointly, and shares hierarchical deep features.
Through this jointly learning strategy, weakly supervised object classification and discovery are beneficial to each other.
We test the proposed method on the challenging PASCAL VOC datasets.
The results show that our method can obtain state-of-the-art performance on object classification, and very competitive results on object discovery, with faster testing speed than competitors.

\end{abstract}

\begin{keyword}

Patch feature learning \sep Multiple instance learning \sep Weakly supervised learning \sep Convolutional neural network \sep End-to-end \sep Object classification \sep Object discovery

\end{keyword}

\end{frontmatter}

\section{Introduction}
\label{sec:intro}

\begin{figure}[!t]
\centering
\centerline{
\includegraphics[width=0.9\linewidth]{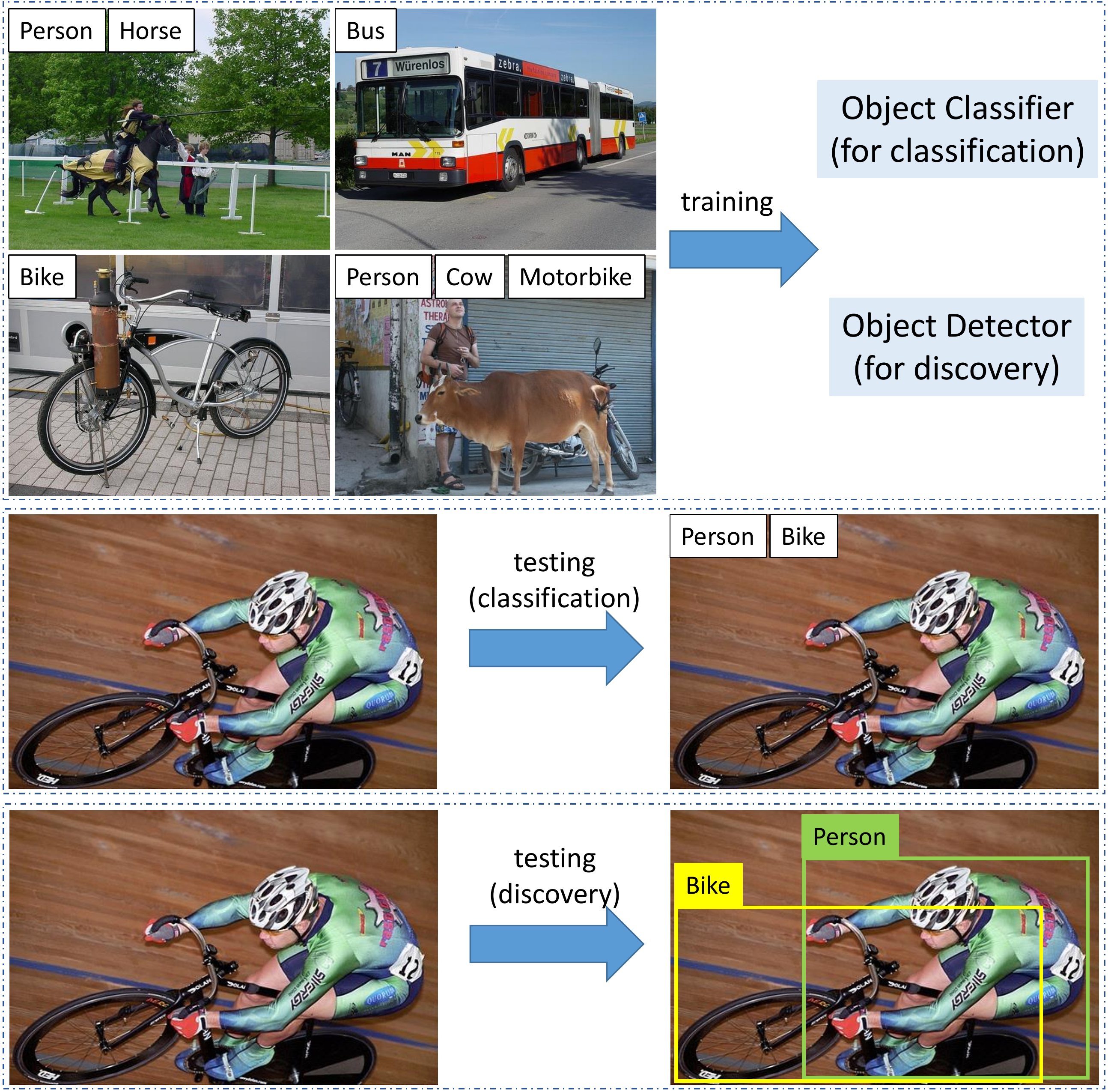}
}
\caption{Illustration of weakly supervised object classification and discovery.
Only image-level annotations are given for training (top).
For classification, object classifier tends to classify which object classes appear in images (middle).
For discovery, object discoverer tends to detect the location of objects in images (bottom).}
\label{fig:task}
\end{figure}

In this paper, we study the problems of weakly supervised object classification and discovery, which are with great importance in computer vision community.
As shown in the top and middle of Fig.~\ref{fig:task}, given an input image and its category labels (e.g., image-level annotations), object classification is to learn object classifiers for classifying which object classes (e.g., person) appear in testing images.\footnote{We refer this task as weakly supervised object classification since it does not require patch-level annotations for training.}
Similar to object detection, object discovery is to learn object detectors for detecting the location of objects in input images, as shown in the bottom of Fig.~\ref{fig:task}.
Different from the fully supervised object detection task that requires exhaustive patch-level/bounding-box annotations for training, object discovery is weakly supervised, i.e., only image-level annotations are necessary to train object discovery models, as shown in the top of Fig.~\ref{fig:task}.\footnote{Object discovery is also called weakly supervised object detection, common object detection, etc., in other papers.}
Nowadays, large scale datasets with patch-level annotations are available \cite{Ref:Deng2009,Ref:Everingham2010,Ref:Lin2014},
and many object classification and detection methods are benefited from such fine-grained annotations \cite{Ref:Girshick2014,Ref:He2015,Ref:Girshick2015,Ref:Yang2016}. However, compared with the great amounts of images with only image-level annotations (e.g., using image search queries to search on the Internet), the amount of exhaustively annotated images is still relatively small.
This inspires us to explore methods that can deal with only image-level annotations.

A popular solution for weakly supervised learning is Multiple Instance Learning (MIL) \cite{Ref:Dietterich1997}.
In MIL, a set of bags and bag labels are given, and each bag consists of a collection of instances, where instances labels are unknown for training.
MIL has two constraints:
1) if a bag is positive, at least one instance in the bag should be positive;
2) if a bag is negative, all instances in the bag should be negative.
It is natural to treat images as bags and patches as instances.
In addition, patch-level feature has wide applications in computer vision community, like image classification \cite{Ref:Csurka2004,Ref:Yang2009,Ref:Gong2014}, object detection \cite{Ref:Felzenszwalb2010,Ref:Girshick2014}, and object discovery \cite{Ref:Wang2015,Ref:Cinbis2014}.
Then we can combine the patch-level feature with MIL for object classification and discovery.

\begin{figure}[!t]
\centering
\centerline{
\includegraphics[width=0.95\linewidth]{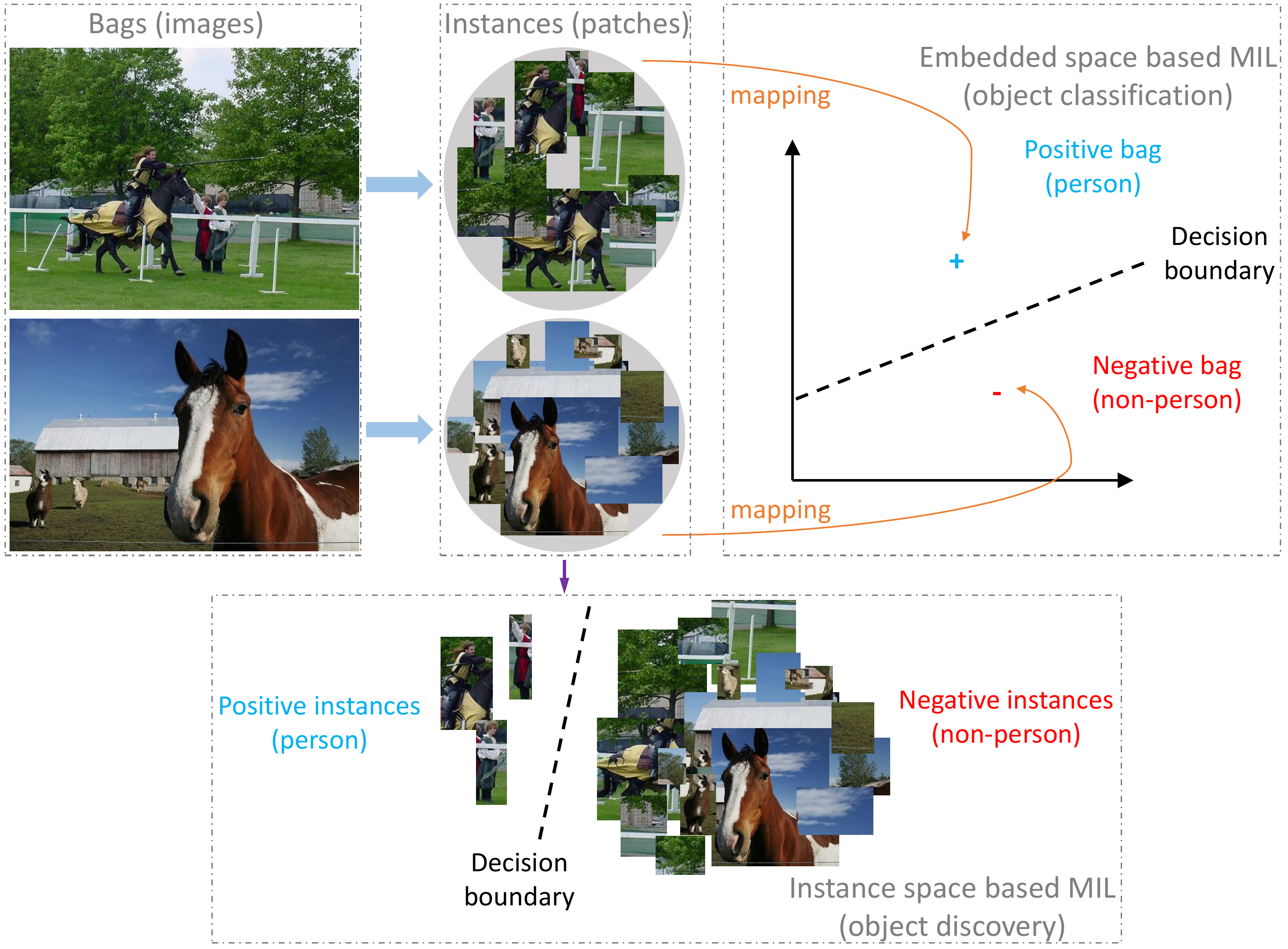}
}
\caption{The relationship among MIL and object classification/discovery.
Images/patches can be viewed as bags/instances.
Embedded space based MIL methods map instances (patches) in each bag (image) to a bag (image) representation for training a bag (image) classifier.
Instance space based MIL methods learn an instance (patch) classifier to discover some most representative instances (patches).
``+''/``-'' indicates positive/negative bag representation.
Note that in the top-right part, only two points are drawn to denote the positive and negative image respectively.}
\label{fig:mil}
\end{figure}

Specifically, the connections between MIL methods and weakly supervised object classification and  discovery are introduced as follows. 
As defined in \cite{Ref:Amores2013}, there are three paradigms for MIL: instance space based MIL methods learn an instance classifier, bag space based MIL methods learn the similarity among bags, and embedded space based MIL methods map bags to representations.
For object classification, many methods aggregate extracted patch features into a vector for each image as image representation, and use the representation to train a classifier \cite{Ref:Csurka2004,Ref:Yang2009,Ref:Cimpoi2015,Ref:Simonyan2015}, which is similar to embedded space based MIL methods as in the top-right of Fig.~\ref{fig:mil}.
Meanwhile for object discovery, instance space based MIL methods are directly applied on patch features to find object of interest \cite{Ref:Wang2015,Ref:Cinbis2014}, as shown in the bottom of Fig.~\ref{fig:mil}.


Recently, deep Convolutional Neural Networks~(CNNs) \cite{Ref:LeCun1989} have obtained great success on image classification \cite{Ref:Krizhevsky2012}.
However, conventional CNNs without patch-level image features are unsuitable to recognize complex images and unable to obtain state-of-the-art performance on challenging datasets, e.g., PASCAL VOC datasets \cite{Ref:Everingham2010}.
There are many reasons: Unlike ImageNet, which have millions object centered images, in PASCAL VOC, (1) There is a limited number of training images; (2) The images have complex structure, and objects have large spatial transformation and scale variation; (3) The images have multiple labels.

Now the state-of-the-art object classification methods for complex datasets are based on local image patches and CNNs \cite{Ref:Wei2015,Ref:Liu2014,Ref:Cimpoi2015,Ref:Simonyan2015}.
And as shown in Fig.~\ref{fig:mil}, it is natural to treat object classification in complex images as a MIL problem.
Thus, it is important to combine deep CNNs with MIL.
There are a few early attempts.
For example, similar to embedded space based MIL methods, Cimpoi et al. \cite{Ref:Cimpoi2015} combines CNN-based patch features with Fisher Vector \cite{Ref:Sanchez2013} to learn image representations.
The Hypotheses CNN Pooling (HCP) \cite{Ref:Wei2015} and Deep Multiple Instance Learning (DMIL) \cite{Ref:Wu2015} find the most representative patches in images.
These examples show that the patch-based CNN has its advantage over plain CNNs.
Also, for object discovery, many methods use CNN to extract patch features, and discover objects by instance space based MIL methods \cite{Ref:Wang2015,Ref:Wang2014}.
All these methods are patch-based, and they are more preferable on complex datasets than plain CNNs.


However, these methods have some limitations.
First, they separately feed each patch into CNN models for feature extraction, ignoring the fact that
computation on convolutional layers for overlapping patches can be shared, thus reduced, in both training and testing procedures.
Second, they treat patch feature extraction, image representation learning, and object classification, or discovery as separate stages.
During training, every stage requires its own training data, taking up a lot of disk space for storage.
At the same time, treating these stages separately may harm performance, as the stages may not be independent.
Therefore it is better to integrate them into a unified framework.
Third, features of patches are extracted using pre-trained models, i.e., they can not learn dataset or task specific patch features.
Last, they treat object classification and discovery as independent tasks, which have been demonstrated to be complementary by our experiments.
Inspired by these facts, we propose a novel framework, called Deep Patch Learning~(DPL), which integrates patch feature learning, image representation learning, object classification and discovery into a unified framework.

Inspired by the fully supervised object detection methods SPPnet \cite{Ref:He2015} and Fast R-CNN \cite{Ref:Girshick2015}, our DPL reduces the training and testing time by sharing the computation on convolutional layers for different patches.
Meanwhile, it combines different stages of object classification and discovery to form an end-to-end framework for classification and discovery.
That is, DPL optimizes the patch feature learning, image representation learning, and image classifying jointly by backpropagation, which mainly focuses on object classification.
In the meantime, it uses a MIL loss for each patch feature, and trains a deep MIL network end-to-end, which can discover the most representative patches in images.
These two blocks (object classification block and MIL based discovery block) are combined via a multi-task learning framework, which boosts the performance for each task.
Moreover, as images may have multiple labels, the MIL loss is adapted to make it suitable for the multi-class case.
Notice that for both object classification and discovery, only image-level annotations are utilized for training, which makes our method quite different from the fully supervised methods \cite{Ref:He2015,Ref:Girshick2015} that require detailed patch-level supervisions.

To demonstrate the effectiveness of our method, we perform elaborate experiments on the PASCAL VOC 2007 and 2012 datasets.
The DPL achieves state-of-the-art performance on object classification, and very competitive results on object discovery.
Moreover, it takes only $1.85$s and $2.8$s for each image during testing, using AlexNet \cite{Ref:Krizhevsky2012} and VGG16 \cite{Ref:Simonyan2015} CNN backend, respectively, which is much faster than the previous best performed method HCP \cite{Ref:Wei2015}.

To summarize, the main contributions of our work are as follows.
\begin{itemize}
\item We propose a weakly supervised learning framework to integrate different stages of object classification into a single deep CNN framework, in order to learn patch features for object classification in an end-to-end manner.
The proposed object classification network is much more effective and efficient than previous patch-based deep CNNs.
\item We novelly integrate the two MIL constraints into the loss of our deep CNN framework to train instance classifiers, which can be applied for object discovery.
\item We embed two tasks object classification and discovery into a single network, and perform classification and discovery simultaneously. We also demonstrate that the two tasks are complementary to some extent.
To the best of our knowledge, it is the first time to demonstrate that classification and discovery can be complementary to each other in an end-to-end neural network. We think this reveals new phenomenon that makes sense to this community.
\item Our method achieves state-of-the-art performance on object classification, and very competitive results on discovery, with faster testing speed on PASCAL VOC datasets.
\end{itemize}

The rest of this paper is organized as follows. In Section~\ref{sec:related_work}, related work is listed. In Section~\ref{sec:DPL}, the detailed architecture of our DPL is described. In Section~\ref{sec:exp}, we present some experiments on several object classification and discovery benchmarks. In Section~\ref{sec:discu}, some discussions of experimental setups are presented. Section~\ref{sec:conclu} concludes the paper.

\section{Related Work}
\label{sec:related_work}

\subsection{MIL and Its Application for Weakly Supervised Object Classification and Discovery}

MIL was first proposed by Dietterich et al. \cite{Ref:Dietterich1997} for drug activity prediction.
Then many methods have emerged in the MIL community \cite{Ref:Wang2015,Ref:Zhang2001,Ref:Andrews2002,Ref:Wei2016}.
Our method can be regarded as a MIL based method as we treat images as bags and patches as instances.
Meanwhile, learning image representations can be viewed as embedded space based MIL method and learning instance classifier can be viewed as instance space based MIL method.
However, traditional MIL methods mainly focus on the problem that bags only have one single label, while in real-world tasks each bag may be associated with more than one class label, e.g., an image may contains multiple objects from different classes.
A solution for the multi-class problem is to adapt the MIL by training a binary classifier for each class through the one-vs.-all strategy \cite{Ref:Christopher2006}.
And the Multi-Instance Multi-Label (MIML) problem \cite{Ref:Zhou2007,Ref:Zhang2008,Ref:Nguyen2010} also have been proposed instead of the single label MIL problem.
As many images in the PASCAL VOC datasets have multiple objects from different classes, our method is also based on the MIML.
Similar to the one-vs.-all strategy, we train some binary classifiers using the multi-class sigmoid cross entropy loss.
But instead of training these binary classifiers separately, we train all classifiers jointly and share features among these classifiers, just like the multi-task learning \cite{Ref:Caruana1997}.
Moreover, different from previous MIL methods, we integrate the MIL constraints into the popular deep CNN, and apply our method to object classification and discovery.

\begin{figure*}[!t]
\centering
\centerline{
\includegraphics[width=0.99\linewidth]{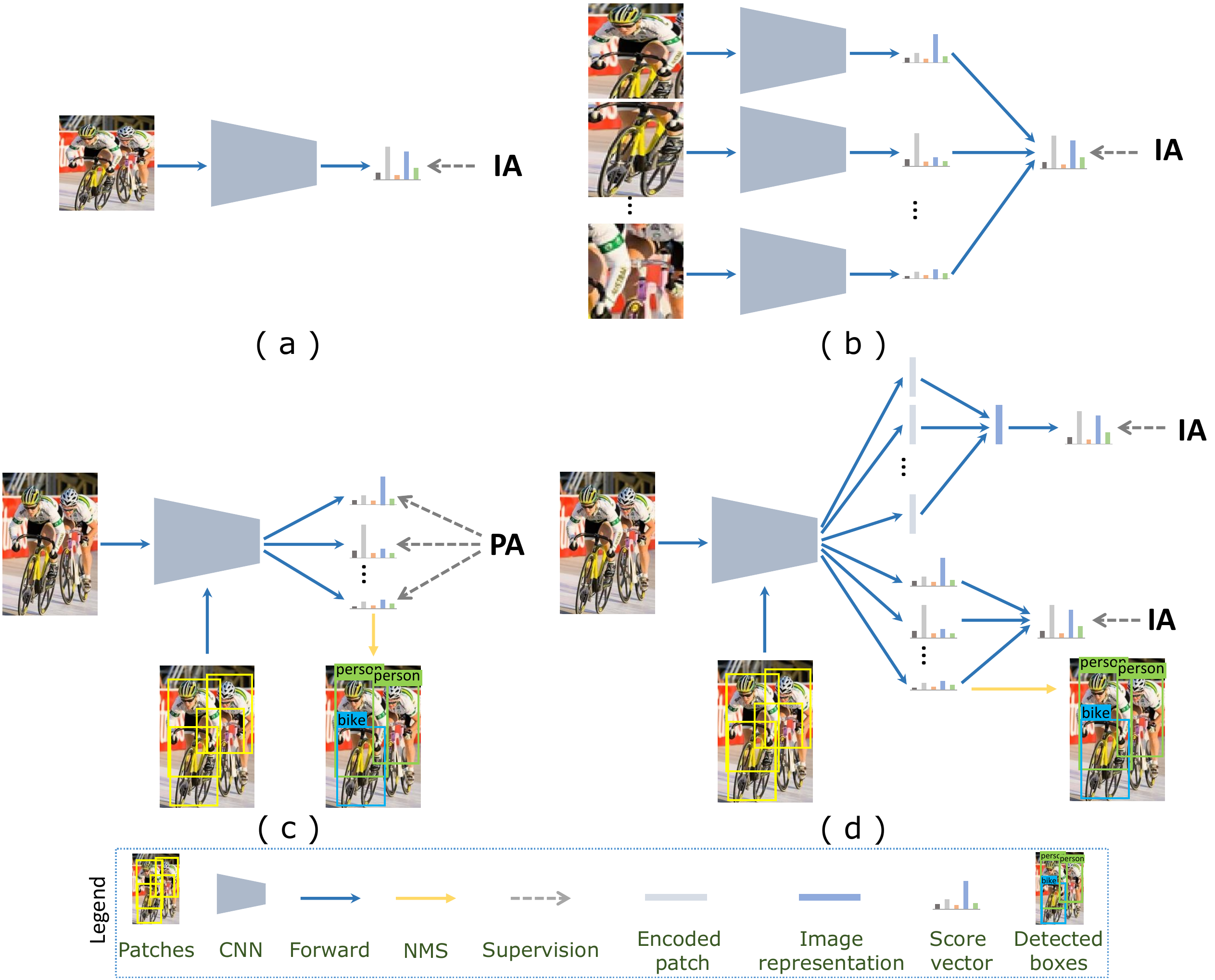}
}
\caption{Illustration of different network architectures:
(a) plain deep CNN;
(b) DMIL \cite{Ref:Wu2015}/HCP \cite{Ref:Wei2015};
(c) Fast R-CNN \cite{Ref:Girshick2015};
(d) our DPL.
Where IA, PA, and NMS mean image-level annotations, patch-level annotations, and non-maxima suppression, respectively.
For (a), a whole image (resized to fixed-size) is fed into the network.
For (b), a set of patches (resized to fixed-size) from one image are fed into the network, and each patch passes a CNN separately.
(a) and (b) produce a score vector per-image for classification and only require image-level annotations for training.
For (c) and (d), a whole image (with original aspect ratio) as well as some patch regions are fed into the network, where all patches from one image share the computation of convolutional layers.
(c) and (d) produce a score vector per-patch, and then NMS is used to filter some highly overlapped patches and produce some detected boxes.
But (c) requires patch-level/bounding-box annotations for training, whereas (d) only takes image-level annotations as supervision.
(d) also produces a score vector per-image for classification.
For simplicity, backpropagation arrows are not plotted in the networks.
Best viewed in color.}
\label{fig:comparision}
\end{figure*}

There are also many other computer vision methods benefit from the MIL.
Wei et al. \cite{Ref:Wei2015} and Wu et al. \cite{Ref:Wu2015} have combined the CNN and MIL for end-to-end object classification.
Their methods are also be end-to-end trainable and can learn patch features.
However, their methods have to resize patches to a special size and feed all patches into the CNN models separately, as shown in Figure~\ref{fig:comparision} (b).
This will result in huge time consumption for training and testing due to ignoring the fact that computation on convolutional layers for overlapping patches can be shared.
Meanwhile, \cite{Ref:Wei2015,Ref:Wu2015} use instance space based MIL methods for solution, which means they train an instance classifier under the MIL constraints, and aggregate instance scores by max-pooling as bag scores. Then they classify bags (images) by these pooled bag scores.
Different from their methods, as shown in Figure~\ref{fig:comparision} (d), we share computation of convolutional layers among different patches, and combine both embedded space and instance space based MIL methods into a single network, which can achieve much promising results.

MIL is also a prevalent method for object discovery.
Cinbis et al. \cite{Ref:Cinbis2014} and Wang et al. \cite{Ref:Wang2015} have used MIL for object discovery, and have achieved some state-of-the-art performance. But their methods separate patch feature extraction and MIL into two separate stages, which may limit their performance.

\subsection{Patch-based Image Classification}

Patch-based methods are popular for image classification as its robustness for spatial transformation, scale variation, and cluttered background.
BoF \cite{Ref:Csurka2004} is a very popular pipeline for image classification. It extracts a set of local features like SIFT \cite{Ref:Lowe2004} or HOG \cite{Ref:Dalal2005} from patches, and then uses some unsupervised ways \cite{Ref:Yang2009,Ref:Sanchez2013,Ref:Bai2015}, or weakly supervised methods \cite{Ref:Pandey2011,Ref:Singh2012,Ref:Wang2013,Ref:Juneja2013,Ref:Doersch2013,Ref:Sun2013,Ref:Shi2016,Ref:Zhou2016,Ref:Tang2016a,Ref:Tang2016b} to aggregate patch features for image representation. These image representations are used for image classification. To consider the spatial layout of images, the Spatial Pyramid Matching~(SPM) \cite{Ref:Lazebnik2006} is employed to enhance the performance. But their pipeline treats patch feature extraction, image representation and classification as independent stages, whereas our method integrates these into a single network and trains the network end-to-end.

Recently, Lobel et al. \cite{Ref:Lobel2013} and Parizi et al. \cite{Ref:Parizi2015} have proposed a method to combine the last two stages, i.e., image representation and classification. They learn patterns of patches and image classifier jointly, and the results show they have improved the performance significantly. Sydorov et al. \cite{Ref:Sydorov2014} have proposed a method to learn the parameters of Fisher Vector and image classifier end-to-end. But as a matter of fact, they do not perform real end-to-end classification. That is, although they can learn the image representation and classifier jointly, they still treat patch feature extraction as an independent part. This will lead to a large consumption of time and space for computing and saving the patch features. Different from their methods, our method achieves real end-to-end learning.

Yang et al. \cite{Ref:Yang2016} also proposes to learn local patch level information for object classification.
They propose a multi-view MIL framework, and chooses the Fisher Vector \cite{Ref:Sanchez2013} to aggregate patch features.
But their method is also not end-to-end, and requires fine-grained bounding-box annotations for training, whereas our method is end-to-end and weakly supervised.


\subsection{Fully Supervised Object Detction}

Inspired by the SPPnet \cite{Ref:He2015} and the great success of CNN \cite{Ref:LeCun1989} for image classification \cite{Ref:Krizhevsky2012}, Girshick \cite{Ref:Girshick2015} have proposed a Fast R-CNN method for fast proposal classification method in fully supervised setting.
Their method can also learn patch features.
Our method follows the path of this work to share computation on convolutional layers among all patches.
But as shown in Figure~\ref{fig:comparision} (c) and (d), the differences between our method and \cite{Ref:Girshick2015} are multi-fold:
1) Fast R-CNN focuses on supervised object detection, whereas the proposed DPL focuses on weakly supervised image classification and object discovery.
2) Fast R-CNN requires bounding-box annotations, whereas DPL only requires image-level annotations. Annotating object bounding-boxes is labor- and time-consuming, whereas image-level annotations are easier to obtain.
3) In summary, Fast R-CNN is a fully supervised object detection framework; DPL is a weakly supervised deep learning framework for joint image classification and object discovery.

\section{The Architecture of Deep Patch Learning}
\label{sec:DPL}

\begin{figure*}[!t]
\centering
\centerline{
\includegraphics[width=0.99\linewidth]{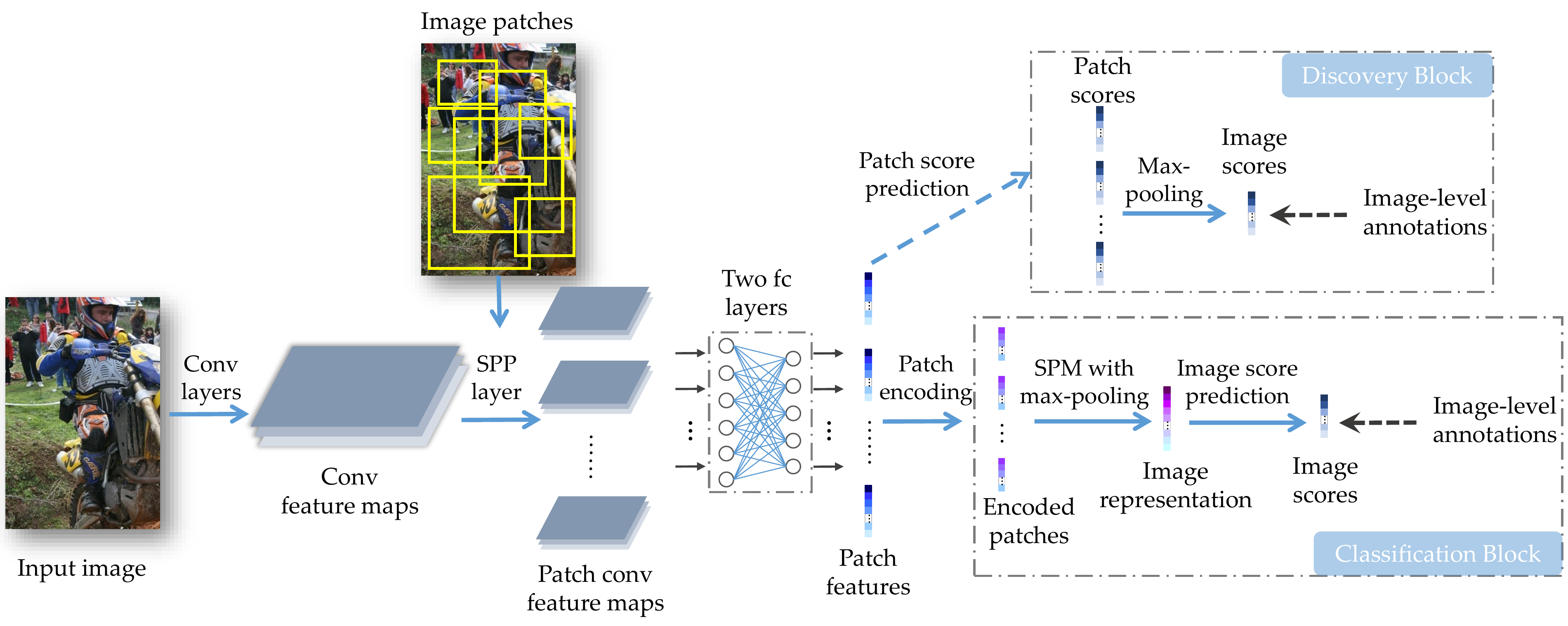}
}
\caption{The architecture of DPL. An image and multiple patches are first input to a fully convolutional network. Each patch is projected to a fixed-size feature map and then fed into several fc layers which will generate a feature vector of each patch. At last, These patch features are branched into two streams: one learns the image representation and classifier jointly for object classification, and the other finds most representative patches for object discovery.
These two streams only require image-level supervisions for training.}
\label{fig:DPL_architecture}
\end{figure*}

The architecture of Deep Patch Learning~(DPL) is shown in Figure~\ref{fig:DPL_architecture}. Given an image and some patches, DPL first passes the image through some convolutional (conv) layers to generate conv feature maps for the whole image, and the size of feature maps is decided by the size of input image. After that, the Spatial Pyramid Pooling~(SPP) layer can be employed for each patch to produce some fixed-size feature maps. Then each feature map can be fed into several fully connected (fc) layers, which will output a set of patch features. At last, these patch features are branched into two different streams with two different tasks: one jointly learns the image representation and classifier focusing on object classification (the classification block), and the other finds most positive patches focusing on object discovery (the discovery block).
Only image-level annotations are used as supervisions to train the two streams.
In this section, we will introduce these steps referred above.

\subsection{Patch Generation}
\label{sec:patch_gen}

Our method is patch-based, so it is necessary to generate patches first.
The simplest and fastest way is sliding window, i.e., sliding a set of fixed-size windows over the image.
But objects only cover a small portion of images and may have various shape, thus patches by fixed-size sliding window are always with low recall.
Some methods propose to generate patches based on some visual cues, like segmentation \cite{Ref:Uijlings2013} and edge \cite{Ref:Zitnick2014}.
Here we choose the ``fast'' mode of Selective Search (SS) \cite{Ref:Uijlings2013} to generate patches due to its fast speed and high recall.

\subsection{Pre-trained CNN Models}
\label{sec:pre_train}

Using CNN models which were trained on large scale datasets like ImageNet \cite{Ref:Deng2009} to fine-tune on target dataset has achieved marvelous performance. We fine-tune our model on two widely used models AlexNet \cite{Ref:Krizhevsky2012} and VGG16 \cite{Ref:Simonyan2015}.

\subsection{CNN and Convolutional Feature Maps}
\label{sec:cnn_map}

As we stated in Section~\ref{sec:pre_train}, we choose two CNN models AlexNet and VGG16.
All the two models have conv layers with some max-pooling layers and three fc layers.
Conv and max-pooling layers are implemented in a sliding window manner.
Actually, all conv and max-pooling layers can deal with inputs of arbitrary sizes, and their outputs maintain roughly the same aspect ratio as input images.
Meanwhile, conv and pooling operations will not change the relative spatial distribution of input images.
Outputs from conv layers are known as conv feature maps \cite{Ref:LeCun1989}.
Though conv and max-pooling layers have the ability to handle arbitrary sized input images, the two CNN models require fix-sized input images because fc layers demand fixed-length input vectors.

\subsection{SPP Layer}

As fc layers take fixed-length input vector, the pre-trained CNN models require fixed-size input image. Therefore, the changing of image size and aspect ratio may somehow leads to loss in the performance. To handle this problem, the Fast R-CNN \cite{Ref:Girshick2015} uses a SPP layer \cite{Ref:He2015} to realize fast proposal classification. Our work follows this path. In special, we replace the last max-pooling layer by the SPP layer.
That is, given $i$-th patch $R_{i}$ and its coordinate is $(l^{x}_{i}, l^{y}_{i}, r^{x}_{i}, r^{y}_{i})$ that indicate the horizontal/vertical ordinates of the top left and bottom right points, suppose the feature maps size is $1/n$ of original image size (e.g., $1/16$ for VGG16), we can project the coordinate of $R_{i}$ to $(l^{x}_{i}/n, l^{y}_{i}/n, r^{x}_{i}/n, r^{y}_{i}/n)$ that corresponds to the coordinate of patch $i$ on feature maps.
Then we can obtain feature maps of patch $i$ by cropping the portion of whole image feature maps inside $R_{i}$ and resizing it to fixed-size.
The size of resized feature maps is depended on the pre-trained CNN model (e.g., $6 \times 6$ for AlexNet and $7 \times 7$ for VGG16).
Taking VGG16 as an example, suppose the $j$-th cropped feature map of $R_{i}$ is $\mathbf{x}_{ij}$, we can divide the $\mathbf{x}_{ij}$ into $7 \times 7$ uniform grids.
Then the output $o^{k}_{ij}$ from the $k$-th grid $R^{k}_{i}$ is as Eq.~(\ref{equ:SPP}).
This procedure will produce fixed-size feature maps for each patch, which can be transmitted to the following fc layers.
More details can be found in \cite{Ref:Girshick2015,Ref:He2015}.


\begin{equation}
\label{equ:SPP}
    o^{k}_{ij} = \mathop{\max} \limits_{\textup{all} R^{k}_{i}} \mathbf{x}_{ij}.
\end{equation}

\subsection{Multi-task Learning Loss}

As shown in Figure~\ref{fig:DPL_architecture} and referred above, our DPL will produce two different scores for two different tasks respectively, one for object classification, and the other for object discovery. Therefore, we replace the last fc layer and the softmax layer of pre-trained models by our multi-task loss. Here we denote the classification loss as $L_{cls}$ and discovery loss as $L_{dis}$, and the total loss is as follows.

\begin{equation}
\label{equ:loss_all}
    L(X_{i}, Y_{i}) = L_{cls}(X_{i}, Y_{i}) + L_{dis}(X_{i}, Y_{i}),
\end{equation}
where $X_{i}$, $Y_{i}$ are the input image and its image-level label respectively. Here we will introduce these two losses in detail.

\subsubsection{Object Classification}

For the object classification, we choose the embedded space based MIL methods.
That is, we learn an image (bag) representation for each image (bag) for classification.
As shown in the classification block of Figure~\ref{fig:DPL_architecture}, after computing the feature vector of each patch, we first encode these patch features. As we all know, an object is composed of a set of parts (e.g., the face and body for person). Patch encoding is to project patch features to more semantic vectors whose elements correspond to patterns (i.e., parts). This can be done by a weight matrix $\mathbf{W} = [\mathbf{w}_{1}, \mathbf{w}_{2}, ..., \mathbf{w}_{N}] \in \mathbb{R}^{K \times N}$, where $i$-th column $\mathbf{w}_{i} \in \mathbb{R}^{K \times 1}$ is the $i$-th weight filter (i.e., $i$-th part filter), $K$ is the dimension of patch features and part filters, $N$ is the number of filters. Suppose the $j$-th patch feature  of image $X_{i}$ is $\mathbf{f}_{ij} \in \mathbb{R}^{K \times 1}$, the encoded patch is represented as $\mathbf{E}_{ij} = [E_{ij1}, E_{ij2}, ..., E_{ijN}]^{T} =  \mathbf{W}^{T} \mathbf{f}_{ij} \in \mathbb{R}^{N \times 1}$.

Then it is necessary to aggregate these encoded patches into an image representation. This can be simply performed by SPM \cite{Ref:Lazebnik2006} with max-pooling. Suppose the SPM split an image into $M$ different grids (i.e., $M$ different sub-regions), then we can aggregate patches in grid $m \in {1, 2, ..., M}$ using max-pooling as $\mathbf{F}_{i}^{m} = [F_{i1}^{m}, F_{i2}^{m}, ..., F_{iN}^{m}]^{T} \in \mathbb{R}^{N \times 1}$, where the element $F_{in}^{m} = \mathop{\max} \limits_{j \in m} E_{ijn}$. And the final image representation is the concatenation of these vectors $\mathbf{F}_{i} = [{\mathbf{F}_{i}^{1}}^{T}, {\mathbf{F}_{i}^{2}}^{T}, ..., {\mathbf{F}_{i}^{M}}^{T}]^{T} \in \mathbb{R}^{NM \times 1}$.

To classify an image, it is necessary to compute the predicted score vector $\mathbf{s}_{cls\_i} \in \mathbb{R}^{C \times 1}$ of the image $X_{i}$ for different classes, where $C$ is the number of classes. Let the classifier be $\mathbf{U}_{cls} = [\mathbf{u}_{cls\_1}, \mathbf{u}_{cls\_2}, ..., \mathbf{u}_{cls\_C}] \in \mathbb{R}^{NM \times C}$, the score can be computed by $\mathbf{s}_{cls\_i} = [s_{cls\_i1}, s_{cls\_i2}, ..., s_{cls\_iC}]^{T} = \mathbf{U}_{cls}^{T} \mathbf{F}_{i} \in \mathbb{R}^{C \times 1}$. Then the loss is as the form of $L_{cls}(X_{i}, Y_{i}) = G(\mathbf{s}_{cls\_i}, Y_{i})$, where $G(\cdot, \cdot)$ is the loss function.

\paragraph{Backpropagation}

To learn the parameters of part filters $\mathbf{W}$ and image classifier $\mathbf{U}_{cls}$, the derivative $\partial L_{cls} / \partial \mathbf{U}_{cls}$ and $\partial L_{cls} / \partial \mathbf{W}$ is required to be computed. This can be easily achieved by standard backpropagation, as shown in Eq.~(\ref{equ:der_u_cls}) and Eq.~(\ref{equ:der_w}).

\begin{equation}
\label{equ:der_u_cls}
    \frac{\partial L_{cls}}{\partial \mathbf{U}_{cls}} = \frac{1}{I} \mathop{\sum} \limits_{i=1}^{I} \mathop{\sum} \limits_{c=1}^{C} \frac{\partial G(\mathbf{s}_{cls\_i}, Y_{i})}{\partial s_{cls\_ic}} \frac{\partial s_{cls\_ic}}{\partial \mathbf{U}_{cls}},
\end{equation}

\begin{equation}
\label{equ:der_w}
    \frac{\partial L_{cls}}{\partial \mathbf{W}} = \frac{1}{I} \mathop{\sum} \limits_{i=1}^{I} \mathop{\sum} \limits_{c=1}^{C} \mathop{\sum} \limits_{r=1}^{NM} \mathop{\sum} \limits_{j=1}^{J_{i}} \mathop{\sum} \limits_{n=1}^{N} \frac{\partial G(\mathbf{s}_{cls\_i}, Y_{i})}{\partial s_{cls\_ic}} \frac{\partial s_{cls\_ic}}{\partial F_{ir}} \frac{\partial F_{ir}}{\partial E_{ijn}} \frac{\partial E_{ijn}}{\partial \mathbf{W}},
\end{equation}
where $I$ is the batch size per-iteration and $J_{i}$ is the patch number of image $X_{i}$. Actually the connections between patch features and encoded patches, image representation and predicted scores are the matrix multiplication, which can be performed by fc layer and is a standard layer in CNN, so we do not give the detailed derivatives of $\partial s_{cls\_ic} / \mathbf{U}_{cls}$, $\partial s_{cls\_ic} / \partial F_{ir}$, and $\partial E_{ijn} / \partial \mathbf{W}$. The derivative of the SPM with max-pooling layer is computed by

\begin{equation}
\label{equ:der_spm}
    \frac{\partial F_{ir}}{\partial E_{ijn}} =
    \begin{cases}
        1 \ \ &\textup{if} \  (r \  \textup{mod} \  N) = n \ \&\& \ j = \mathop{arg\,max} \limits_{j' \in m} E_{ij'n},\\
        0 \ \ &\textup{otherwise}.
    \end{cases}
\end{equation}
Where the $\textup{mod}$ is the operation that computes the remainder, and $m$ is the $m$-th grid satisfying $m = ceil(r / N)$. Through the backpropagation, an end-to-end system for patch feature learning, image representation, and classification can be obtained.

\subsubsection{Object Discovery}

Different from the object classification, which aims at finding some important parts to compose the object, the object discovery is to find the patch that can locate the object exactly. That is, object classification tends to learn the local information of an object, and object discovery tends to learn the global information of an object. The two tasks are complementary in some degree, so here we also perform object discovery, as shown in the discovery block of Figure~\ref{fig:DPL_architecture}.

Object discovery and instance space based MIL method have similar targets. That is, the former wants to find the most representative patches of an object in the image, and the latter wants to find positive instances in the positive bag. If we treat image as bag and patches as instances, these two concepts may be equivalent. There is other work that utilizes instance space based MIL methods to realize object discovery \cite{Ref:Wang2015,Ref:Cinbis2014}. Our object discovery method also adopts this method to find the most positive patch of the object, just as the MI-SVM \cite{Ref:Andrews2002}.

Specially, we define the patch classifier $\mathbf{U}_{dis} = [\mathbf{u}_{dis\_1}, \mathbf{u}_{dis\_2}, ..., \mathbf{u}_{dis\_C}] \in \mathbb{R}^{K \times C}$. Then the scores of patch feature $\mathbf{f}_{ij}$ can be computed by $\mathbf{s}_{pat\_ij} = [s_{pat\_ij1}, s_{pat\_ij2}, ..., s_{pat\_ijC}]^{T} = \mathbf{U}_{dis}^{T} \mathbf{f}_{ij} \in \mathbb{R}^{C \times 1}$. As in the MIL constraints, there must be at least one positive instance in a positive bag, and all instances should be negative in negative bags. So if an image contains an object, the maximum score of patches corresponding to that object should be a large value, but if an image does not contain an object, the maximum score of patches corresponding to that object should be a small value. This can be realized by the max-pooling over all patches, that is, $s_{dis\_ic} = \mathop{\max} \limits_{j} s_{pat\_ijc}$, where $s_{dis\_ic}$ is the score that indicates whether image $X_{i}$ contains $c$-th object. So we can define $\mathbf{s}_{dis\_i} = [s_{dis\_i1}, s_{dis\_i2}, ..., s_{dis\_iC}]^{T} \in \mathbb{R}^{C \times 1}$ as the predicted score vector for the image, which aims at object discovery. The $\mathbf{s}_{dis\_i}$ and $\mathbf{s}_{cls\_i}$ are similar, and both of them represent the predict score vector of an image, so we can use the same loss function $L_{dis}(X_{i}, Y_{i}) = G(\mathbf{s}_{dis\_i}, Y_{i})$.

\paragraph{Backpropagation}

To learn the parameters of patch classifier $\mathbf{U}_{dis}$, the derivative $\partial L_{dis} / \partial \mathbf{U}_{dis}$ is required to be computed, which can be easily achieved by the backpropagation, as shown in Eq.~(\ref{equ:der_u_dis}).

\begin{equation}
\label{equ:der_u_dis}
    \frac{\partial L_{dis}}{\partial \mathbf{U}_{dis}} = \frac{1}{I} \mathop{\sum} \limits_{i=1}^{I} \mathop{\sum} \limits_{c=1}^{C} \mathop{\sum} \limits_{j=1}^{J_{i}} \frac{\partial G(\mathbf{s}_{dis\_i}, Y_{i})}{\partial s_{dis\_ic}} \frac{\partial s_{dis\_ic}}{\partial s_{pat\_ijc}} \frac{\partial s_{pat\_ijc}}{\partial \mathbf{U}_{dis}},
\end{equation}

The connection between patch features and patch scores can also be achieved by the fc layer, so we only give the derivative of the max-pooling layer as Eq.~(\ref{equ:der_max}).

\begin{equation}
\label{equ:der_max}
    \frac{\partial s_{dis\_ic}}{\partial s_{pat\_ijc}} =
    \begin{cases}
        1 \ \ & \textup{if} \ j = \mathop{arg\,max} \limits_{j'} s_{pat\_ij'c}, \\
        0 \ \ & \textup{otherwise}.
    \end{cases}
\end{equation}
Through the backpropagation, end-to-end object discovery can thus be performed.

\subsubsection{Loss}

In the above two subsections, we have derived the backpropagation of object classification and discovery. In this part, we will explain the $G(\mathbf{s}_{i}, Y_{i})$ and its derivative. As one image may contain multiple objects from different classes, so it has multiple labels, and then its label will become a binary label vector $\mathbf{Y}_{i} = [y_{i1}, y_{i2}, ..., y_{iC}]^{T} \in \mathbb{R}^{C \times 1}$, where $y_{ic} = 1$ if $X_{i}$ has object $c$, $y_{ic} = 0$ otherwise. The popular softmax loss function is not suitable for this case, so we choose multi-class sigmoid cross entropy loss, as shown in Eq.~(\ref{equ:g}).

\begin{equation}
\label{equ:g}
    G(\mathbf{s}_{i}, \mathbf{Y}_{i}) = -\mathop{\sum} \limits_{c=1}^{C} \{y_{ic}\log\, \sigma (s_{ic}) + (1 - y_{ic})\log\,(1 - \sigma (s_{ic}))\},
\end{equation}
where $\sigma (x)$ is the sigmoid function $\sigma (x) = 1 / (1 + \exp(-x))$.
Using the Eq.~(\ref{equ:g}), we train $C$ binary classifiers each of which distinguishes images are with/without one object class, just similar to the one-vs.-all strategy \cite{Ref:Christopher2006} for multi-class classification.
After that, the derivative of Eq.~(\ref{equ:g}) can be obtained as follows.

\begin{equation}
\label{equ:der_g}
    \frac{\partial G(\mathbf{s}_{i}, \mathbf{Y}_{i})}{\partial s_{ic}} = \sigma (s_{ic}) - y_{ic}.
\end{equation}
Then all the derivatives of parameters can be derived.
We can observe that only image-level labels $\mathbf{Y}_{i}$ are necessary to optimize the loss in Eq.~(\ref{equ:g}), which confirms our method is totally weakly supervised.

\section{Experiments}
\label{sec:exp}

In this section we will show the experiments of our DPL method for object classification and discovery.

\subsection{Experimental Setup}

\subsubsection{Details of the Architecture}

As stated in Section~\ref{sec:pre_train}, we choose two popular CNN architectures AlexNet \cite{Ref:Krizhevsky2012} and VGG16 \cite{Ref:Simonyan2015}, which are pre-trained on the ImageNet \cite{Ref:Deng2009}. These pre-trained models can be downloaded from the Caffe model zoo\footnote{\url{https://github.com/BVLC/caffe/wiki/Model-Zoo}}. We replace the last max-pooling layer, the final fc layer, and the softmax loss layer by the layers defined in Section~\ref{sec:DPL}. The dimension of encoded patch is set to $256$ (i.e., $N=256$). Then we choose three different SPM scales $\{1\times1, 2\times2, 3\times1\}$ for the SPM with max-pooling layer after the patch encoding layer. The fc layers for patch encoding, image and patch score prediction are initialized using Gaussian distributions with $0$-mean and standard deviations $0.01$. Biases are initialized to be $0$. The mini-batch size is set to $2$. For AlexNet, learning rates of all layers are set to $0.001$ in the first $30$K mini-batch iterations and $0.0001$ in the later $10$K iterations. For VGG16, as it is very deep and hard to train, we first only train the layers after the second fc layer $5$K iterations with learning rate $0.001$, and then train another $40$K iterations as for AlexNet. The momentum and weight decay are set to $0.9$ and $0.0005$ respectively.

\begin{table}[!t]\footnotesize
\begin{center}
\caption{Object classification results (AP in $\%$) for different methods on PASCAL VOC 2007 test set.}
\label{table:cls_2007}
\resizebox{\linewidth}{!}{
\begin{tabular}{@{}L{3.2cm}|*{10}{x}|x}
\toprule
method & aero & bike & bird & boat & bottle & bus & car & cat & chair & cow &\\
\midrule
DMIL \cite{Ref:Wu2015} & $93.5$ & $83.4$ & $86.9$ & $83.6$ & $54.2$ & $81.6$ & $86.6$ & $85.2$ & $54.5$ & $68.9$\\
CNNaug-SVM \cite{Ref:Razavian2014} & $90.1$ & $84.4$ & $86.5$ & $84.1$ & $48.4$ & $73.4$ & $86.7$ & $85.4$ & $61.3$ & $67.6$\\
Oquab et al. \cite{Ref:Oquab2014} & $88.5$ & $81.5$ & $87.9$ & $82.0$ & $47.5$ & $75.5$ & $90.1$ & $87.2$ & $61.6$ & $75.7$ &\\
Chatfield et al. \cite{Ref:Chatfield2014} & $95.3$ & $90.4$ & $92.5$ & $89.6$ & $54.4$ & $81.9$ & $91.5$ & $91.9$ & $64.1$ & $76.3$ &\\
Barat et al. \cite{Ref:Barat2016} & - & - & - & - & - & - & - & - & - & - &\\
HCP-AlexNet \cite{Ref:Wei2015} & $95.4$ & $90.7$ & $92.9$ & $88.9$ & $53.9$ & $81.9$ & $91.8$ & $92.6$ & $60.3$ & $79.3$ &\\
Cimpoi et al. \cite{Ref:Cimpoi2015} & - & - & - & - & - & - & - & - & - & - &\\
VGG16-SVM \cite{Ref:Simonyan2015} & - & - & - & - & - & - & - & - & - & - &\\
VGG19-SVM \cite{Ref:Simonyan2015} & - & - & - & - & - & - & - & - & - & - &\\
VGG16-19-SVM \cite{Ref:Simonyan2015} & - & - & - & - & - & - & - & - & - & - &\\
FeV+LV-20 \cite{Ref:Yang2016} & $97.9$ & $97.0$ & $96.6$ & $94.6$ & $73.6$ & $93.9$ & $96.5$ & $95.5$ & $73.7$ & $90.3$&\\
HCP-VGG16 \cite{Ref:Wei2015} & $\mathbf{98.6}$ & $97.1$ & $\mathbf{98.0}$ & $95.6$ & $75.3$ & $\mathbf{94.7}$ & $95.8$ & $97.3$ & $73.1$ & $90.2$ &\\
\midrule
DPL-AlexNet & $94.6$ & $92.0$ & $90.9$ & $88.8$ & $61.8$ & $85.4$ & $95.0$ & $91.8$ & $68.8$ & $77.1$ &\\
DPL-VGG16 & $98.0$ & $\mathbf{97.4}$ & $96.8$ & $\mathbf{95.6}$ & $\mathbf{81.3}$ & $94.6$ & $\mathbf{97.7}$ & $\mathbf{97.7}$ & $\mathbf{79.7}$ & $\mathbf{91.1}$ &\\
\bottomrule
\toprule
method & table & dog & horse & mbike & persn & plant & sheep & sofa & train & tv & mAP\\
\midrule
DMIL \cite{Ref:Wu2015} & $53.8$ & $73.2$ & $78.8$ & $79.0$ & $86.6$ & $51.2$ & $74.4$ & $63.7$ & $91.5$ & $80.4$ & $75.5$\\
CNNaug-SVM \cite{Ref:Razavian2014} & $69.6$ & $84.0$ & $85.4$ & $80.0$ & $92.0$ & $56.9$ & $76.7$ & $67.3$ & $89.1$ & $74.9$ & $77.2$\\
Oquab et al. \cite{Ref:Oquab2014} & $67.3$ & $85.5$ & $83.5$ & $80.0$ & $95.6$ & $60.8$ & $76.8$ & $58.0$ & $90.4$ & $77.9$ & $77.7$\\
Chatfield et al. \cite{Ref:Chatfield2014} & $74.9$ & $89.7$ & $92.2$ & $86.9$ & $95.2$ & $60.7$ & $82.9$ & $68.0$ & $95.5$ & $74.4$ & $82.4$\\
Barat et al. \cite{Ref:Barat2016} & - & - & - & - & - & - & - & - & - & - & $82.5$\\
HCP-AlexNet \cite{Ref:Wei2015} & $73.0$ & $90.8$ & $89.2$ & $86.4$ & $92.5$ & $66.9$ & $86.4$ & $65.6$ & $94.4$ & $80.4$ & $82.7$\\
Cimpoi et al. \cite{Ref:Cimpoi2015} & - & - & - & - & - & - & - & - & - & - & $88.6$\\
VGG16-SVM \cite{Ref:Simonyan2015} & - & - & - & - & - & - & - & - & - & - & $89.3$\\
VGG19-SVM \cite{Ref:Simonyan2015} & - & - & - & - & - & - & - & - & - & - & $89.3$\\
VGG16-19-SVM \cite{Ref:Simonyan2015} & - & - & - & - & - & - & - & - & - & - & $89.7$\\
FeV-LV-20 \cite{Ref:Yang2016} & $82.8$ & $95.4$ & $\mathbf{97.7}$ & $\mathbf{95.9}$ & $98.6$ & $77.6$ & $88.7$ & $78.0$ & $\mathbf{98.3}$ & $89.0$ & $90.6$\\
HCP-VGG16 \cite{Ref:Wei2015} & $80.0$ & $\mathbf{97.3}$ & $96.1$ & $94.9$ & $96.3$ & $78.3$ & $\mathbf{94.7}$ & $76.2$ & $97.9$ & $91.5$ & $90.9$\\
\midrule
DPL-AlexNet & $77.6$ & $88.5$ & $92.7$ & $90.3$ & $98.0$ & $74.3$ & $84.8$ & $75.7$ & $95.5$ & $82.6$ & $85.3$\\
DPL-VGG16 & $\mathbf{85.7}$ & $96.5$ & $96.9$ & $95.8$ & $\mathbf{99.2}$ & $\mathbf{83.1}$ & $92.2$ & $\mathbf{83.8}$ & $97.8$ & $\mathbf{92.8}$ & $\mathbf{92.7}$\\
\bottomrule
\end{tabular}
}
\end{center}
\end{table}

\subsubsection{Datasets and Evaluation Measures}

We test our DPL method on two famous object classification and discovery benchmarks PASCAL VOC 2007 and PASCAL VOC 2012 \cite{Ref:Everingham2010}, which have $9,962$ and $22,531$ images respectively with 20 different object categories. The datasets are split into standard train, val and test sets. We use the trainval set ($5,011$ images for VOC 2007 and $11,540$ images for VOC 2012) with only image-level labels to train our models. During the testing procedure, for object classification, we compute Average Percision~(AP) and the mean of AP~(mAP) as the evaluation metric to test our model on the test set\footnote{For VOC 2012, the evaluation is performed online via the PASCAL VOC evaluation server (\url{http://host.robots.ox.ac.uk:8080/}).} ($4,952$ images for VOC 2007 and $10,991$ images for VOC 2012). For object discovery, we report the CorLoc on the trainval set as in \cite{Ref:Deselaers2012}, which computes the percentage of the correct location of objects under the PASCAL criteria (Intersection over Union~(IoU) $>0.5$ between the ground truths and predicted bounding boxes).

\subsubsection{Patch Generation Protocols}

There are many different methods to generate patches, like proposal based methods Selective Search~(SS) \cite{Ref:Uijlings2013} and EdgeBoxes \cite{Ref:Zitnick2014}, or sliding widow, which need $1.5$s, $0.25$s, and less than $0.01$s respectively (we use the ``fast'' mode of SS). To get patches, we choose SS \cite{Ref:Uijlings2013} to produce $1$-$3$K patches for each image. For data augmentation, we use five image scales $\{480, 576, 688, 864, 1200\}$ (resize the longest side to one of the scales and maintain the aspect ratio of images) with their horizontal flips to train the model. For testing, we use the same five scales without flips and compute the mean score of these scales.

\begin{table}[!t]\footnotesize
\begin{center}
\caption{Object classification results (AP in $\%$) for different methods on PASCAL VOC 2012 test set.}
\label{table:cls_2012}
\resizebox{\linewidth}{!}{
\begin{tabular}{@{}L{3.2cm}|*{10}{x}|x}
\toprule
method & aero & bike & bird & boat & bottle & bus & car & cat & chair & cow &\\
\midrule
DDSFL \cite{Ref:Zuo2015} & $92.7$ & $75.4$ & $85.1$ & $79.4$ & $42.3$ & $88.1$ & $68.5$ & $87.1$ & $62.9$ & $65.8$ &\\
HCP-AlexNet \cite{Ref:Wei2015} & $97.7$ & $83.2$ & $92.8$ & $88.5$ & $60.1$ & $88.7$ & $82.7$ & $94.4$ & $65.8$ & $81.9$ &\\
Oquab et al. \cite{Ref:Oquab2014} & $94.6$ & $82.9$ & $88.2$ & $84.1$ & $60.3$ & $89.0$ & $84.4$ & $90.7$ & $72.1$ & $86.8$ &\\
Chatfield et al. \cite{Ref:Chatfield2014} & $96.8$ & $82.5$ & $91.5$ & $88.1$ & $62.1$ & $88.3$ & $81.9$ & $94.8$ & $70.3$ & $80.2$ &\\
Oquab et al. \cite{Ref:Oquab2015} & $96.7$ & $88.8$ & $92.0$ & $87.4$ & $64.7$ & $91.1$ & $87.4$ & $94.4$ & $74.9$ & $89.2$ &\\
VGG16-SVM \cite{Ref:Simonyan2015} & $99.0$ & $88.8$ & $95.9$ & $93.8$ & $73.1$ & $92.1$ & $85.1$ & $97.8$ & $79.5$ & $91.1$ &\\
VGG19-SVM \cite{Ref:Simonyan2015} & $99.1$ & $88.7$ & $95.7$ & $93.9$ & $73.1$ & $92.1$ & $84.8$ & $97.7$ & $79.1$ & $90.7$ &\\
VGG16-19-SVM \cite{Ref:Simonyan2015} & $99.1$ & $89.1$ & $96.0$ & $94.1$ & $74.1$ & $92.2$ & $85.3$ & $97.9$ & $79.9$ & $92.0$ &\\
FeV+LV-20 \cite{Ref:Yang2016} & $98.4$ & $92.8$ & $93.4$ & $90.7$ & $74.9$ & $93.2$ & $90.2$ & $96.1$ & $78.2$ & $89.8$ & \\
HCP-VGG16 \cite{Ref:Wei2015} & $99.1$ & $92.8$ & $\mathbf{97.4}$ & $\mathbf{94.4}$ & $79.9$ & $93.6$ & $89.8$ & $\mathbf{98.2}$ & $78.2$ & $\mathbf{94.9}$ &\\
\midrule
DPL-AlexNet & $96.6$ & $87.1$ & $91.4$ & $89.0$ & $61.8$ & $89.7$ & $88.6$ & $93.9$ & $71.4$ & $77.1$ &\\
DPL-VGG16 & $\mathbf{99.1}$ & $\mathbf{95.0}$ & $96.6$ & $94.1$ & $\mathbf{83.2}$ & $\mathbf{95.0}$ & $\mathbf{94.6}$ & $98.1$ & $\mathbf{82.2}$ & $93.9$ &\\
\bottomrule
\toprule
method & table & dog & horse & mbike & persn & plant & sheep & sofa & train & tv & mAP\\
\midrule
DDSFL \cite{Ref:Zuo2015} & $60.6$ & $80.2$ & $74.0$ & $76.4$ & $92.3$ & $43.6$ & $74.3$ & $53.2$ & $88.4$ & $73.7$ & $73.2$\\
HCP-AlexNet \cite{Ref:Wei2015} & $68.0$ & $92.6$ & $89.1$ & $87.6$ & $92.1$ & $58.0$ & $86.6$ & $55.5$ & $92.5$ & $77.6$ & $81.8$\\
Oquab et al. \cite{Ref:Oquab2014} & $69.0$ & $92.1$ & $93.4$ & $88.6$ & $96.1$ & $64.3$ & $86.6$ & $62.3$ & $91.1$ & $79.8$ & $82.8$\\
Chatfield et al. \cite{Ref:Chatfield2014} & $76.2$ & $92.9$ & $90.3$ & $89.3$ & $95.2$ & $57.4$ & $83.6$ & $66.4$ & $93.5$ & $81.9$ & $83.2$\\
Oquab et al. \cite{Ref:Oquab2015} & $76.3$ & $93.7$ & $95.2$ & $91.1$ & $97.6$ & $66.2$ & $91.2$ & $70.0$ & $94.5$ & $83.7$ & $86.3$\\
VGG16-SVM \cite{Ref:Simonyan2015} & $83.3$ & $97.2$ & $96.3$ & $94.5$ & $96.9$ & $63.1$ & $93.4$ & $75.0$ & $97.1$ & $87.1$ & $89.0$\\
VGG19-SVM \cite{Ref:Simonyan2015} & $83.2$ & $97.3$ & $96.2$ & $94.3$ & $96.9$ & $63.4$ & $93.2$ & $74.6$ & $97.3$ & $87.9$ & $89.0$\\
VGG16-19-SVM \cite{Ref:Simonyan2015} & $\mathbf{83.7}$ & $97.5$ & $96.5$ & $94.7$ & $97.1$ & $63.7$ & $93.6$ & $75.2$ & $\mathbf{97.4}$ & $87.8$ & $89.3$\\
FeV+LV-20 \cite{Ref:Yang2016} & $80.6$ & $95.7$ & $96.1$ & $95.3$ & $97.5$ & $73.1$ & $91.2$ & $75.4$ & $97.0$ & $88.2$ & $89.4$\\
HCP-VGG16 \cite{Ref:Wei2015} & $79.8$ & $\mathbf{97.8}$ & $\mathbf{97.0}$ & $93.8$ & $96.4$ & $74.3$ & $\mathbf{94.7}$ & $71.9$ & $96.7$ & $88.6$ & $90.5$\\
\midrule
DPL-AlexNet & $74.5$ & $91.7$ & $90.5$ & $91.8$ & $97.6$ & $69.8$ & $82.7$ & $67.2$ & $93.2$ & $83.2$ & $84.4$\\
DPL-VGG16 & $82.1$ & $97.6$ & $96.8$ & $\mathbf{96.7}$ & $\mathbf{98.9}$ & $\mathbf{84.1}$ & $93.5$ & $\mathbf{79.4}$ & $97.3$ & $\mathbf{91.3}$ & $\mathbf{92.5}$\\
\bottomrule
\end{tabular}
}
\end{center}
\end{table}

\subsubsection{Experimental Platform}

Our code is written by C++ and Python, based on the Caffe \cite{Ref:Jia2014} and the publicly available implementation of Fast R-CNN \cite{Ref:Girshick2015}. All of our experiments are running on a NVIDIA GTX TitanX GPU with $12$GB memory.
Codes for reproducing the results are available at \url{https://github.com/ppengtang/dpl}.

\begin{figure*}[t]
\centering
\centerline{
\includegraphics[width=0.95\linewidth]{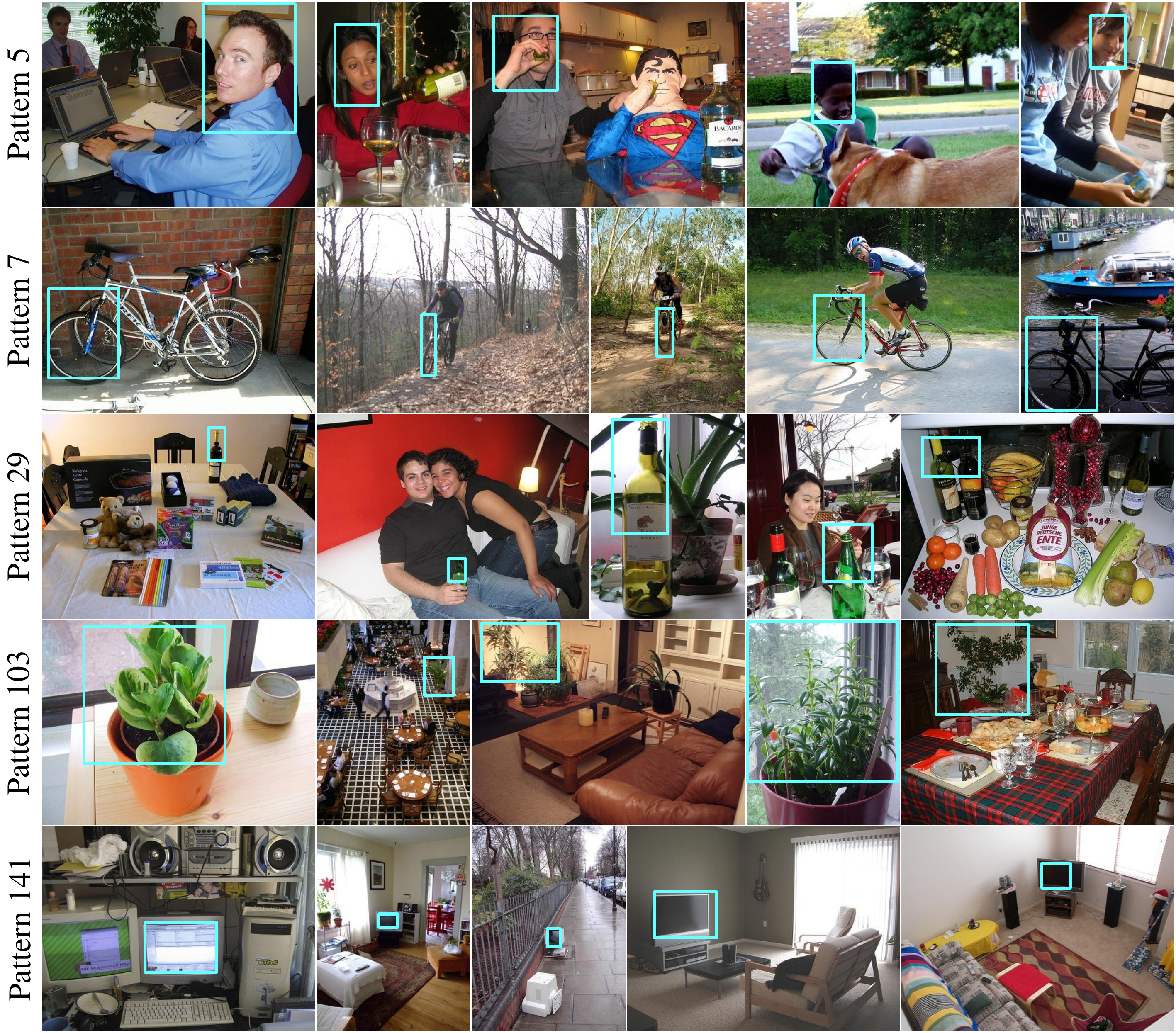}
}
\caption{Visualization of the patterns by our patch encoding method.
Each row corresponds to one patterns, where blue rectangles show patches with strongest responses of the patterns.}
\label{fig:patterns}
\end{figure*}

\begin{table}[!t]\footnotesize
\begin{center}
\caption{Object discovery results (CorLoc in $\%$) for different methods on PASCAL VOC 2007 trainval set.}
\label{table:dis_2007}
\resizebox{\linewidth}{!}{
\begin{tabular}{@{}L{3.2cm}|*{10}{x}|x}
\toprule
method & aero & bike & bird & boat & bottle & bus & car & cat & chair & cow &   \\
\midrule
Shi et al. \cite{Ref:Shi2013} & $67.3$ & $54.4$ & $34.3$ & $17.8$ & $1.3$ & $46.6$ & $60.7$ & $\mathbf{68.9}$ & $2.5$ & $32.4$\\
Multi-fold MIL \cite{Ref:Cinbis2014} & $56.6$ & $58.3$ & $28.4$ & $20.7$ & $6.8$ & $54.9$ & $69.1$ & $20.8$ & $9.2$ & $50.5$\\
RMI-SVM \cite{Ref:Wang2015} & $37.7$ & $58.8$ & $39.0$ & $4.7$ & $4.0$ & $48.4$ & $70.0$ & $63.7$ & $9.0$ & $54.2$ &\\
Bilen et al. \cite{Ref:Bilen2015} & $66.4$ & $59.3$ & $42.7$ & $20.4$ & $21.3$ & $63.4$ & $\mathbf{74.3}$ & $59.6$ & $21.1$ & $\mathbf{58.2}$ &\\
Wang et al. \cite{Ref:Wang2014} & $\mathbf{80.1}$ & $63.9$ & $\mathbf{51.5}$ & $14.9$ & $21.0$ & $55.7$ & $74.2$ & $43.5$ & $\mathbf{26.2}$ & $53.4$ &\\
\midrule
DPL-AlexNet & $77.1$ & $\mathbf{65.9}$ & $42.9$ & $19.2$ & $13.7$ & $59.9$ & $72.3$ & $42.4$ & $8.9$ & $54.1$ &\\
DPL-VGG16 & $79.2$ & $42.0$ & $39.0$ & $\mathbf{42.0}$ & $\mathbf{22.5}$ & $\mathbf{67.5}$ & $70.7$ & $45.6$ & $18.9$ & $42.5$ &\\
\bottomrule
\toprule
method & table & dog & horse & mbike & persn & plant & sheep & sofa & train & tv & Ave.\\
\midrule
Shi et al. \cite{Ref:Shi2013} & $16.2$ & $\mathbf{58.9}$ & $51.5$ & $64.6$ & $18.2$ & $3.1$ & $20.9$ & $34.7$ & $63.4$ & $5.9$ & $36.2$\\
Multi-fold MIL \cite{Ref:Cinbis2014} & $10.2$ & $29.0$ & $58.0$ & $64.9$ & $\mathbf{36.7}$ & $18.7$ & $56.5$ & $13.2$ & $54.9$ & $59.4$ & $38.8$\\
RMI-SVM \cite{Ref:Wang2015} & $\mathbf{33.3}$ & $37.4$ & $\mathbf{61.6}$ & $57.6$ & $30.1$ & $31.7$ & $32.4$ & $\mathbf{52.8}$ & $49.0$ & $27.8$ & $40.2$\\
Bilen et al. \cite{Ref:Bilen2015} & $14.0$ & $38.5$ & $49.5$ & $60.0$ & $19.8$ & $39.2$ & $41.7$ & $30.1$ & $50.2$ & $44.1$ & $43.7$\\
Wang et al. \cite{Ref:Wang2014} & $16.3$ & $56.7$ & $58.3$ & $69.5$ & $14.1$ & $38.3$ & $\mathbf{58.8}$ & $47.2$ & $49.1$ & $\mathbf{60.9}$ & $\mathbf{48.5}$\\
\midrule
DPL-AlexNet & $13.7$ & $40.9$ & $38.8$ & $75.9$ & $35.3$ & $\mathbf{42.9}$ & $57.7$ & $29.8$ & $51.7$ & $26.5$ & $43.5$\\
DPL-VGG16 & $24.7$ & $37.4$ & $35.0$ & $\mathbf{80.3}$ & $17.6$ & $32.2$ & $57.7$ & $37.1$ & $\mathbf{63.9}$ & $52.7$ & $45.4$\\
\bottomrule
\end{tabular}
}
\end{center}
\end{table}

\subsection{Object Classification}

We first report our results for object classification. Even though the discovery block in Figure~\ref{fig:DPL_architecture} mainly focuses on object discovery, it can also produce image-level scores. So for object classification, we compute the mean score of two different tasks. The results on VOC 2007 and VOC 2012 are shown in Table~\ref{table:cls_2007} and Table~\ref{table:cls_2012}.\footnote{The results of our method on VOC 2012 are also available on \url{http://host.robots.ox.ac.uk:8080/anonymous/PRKWXL.html} and \url{http://host.robots.ox.ac.uk:8080/anonymous/PWADSM.html}.}

From the results, we can observe that our method outperforms other CNN-based methods using single model. Specially, our method is better than other patch-based methods for quite a lot. For example, the method in \cite{Ref:Cimpoi2015} extract ten different scale patch features from pre-trained VGG19 model with Fisher Vector. In \cite{Ref:Simonyan2015}, patch features are extracted from five different scales with mean-pooling. HCP \cite{Ref:Wei2015} combines the MIL constraints and CNN models to find the most representative patches in images.
Our method even outperforms the FeV+LV-20 \cite{Ref:Yang2016} that utilizes bounding-box annotations during training, which shows the potential for combining CNNs with weakly supervised methods (e.g., MIL).
As shown in Table~\ref{table:cls_2007} and Table~\ref{table:cls_2012}, our method achieves $1.8\%$ and $2.0\%$ incresement on VOC 2007 and VOC 2012 respectively. These results show that our DPL method can achieve the state-of-the-art performance on object classification. On VOC 2012, the best result was reported in literature is the combination of HCP-VGG16 \cite{Ref:Wei2015} and NUS-PSL \cite{Ref:Yan2012}, which achieves $93.2\%$ mAP, but it just simply averages the predicted scores by two methods.

Some patterns from our patch encoding method are also visualized in Figure~\ref{fig:patterns}.
We can observe that, though only image-level annotations are avaliable during training, our method can learn patterns with great semantic information.
For example, ``Pattern 5'' corresponds to head of person; ``Pattern 7'' corresponds to wheel of bicycle; ``Pattern 141'' corresponds to screen of tvmonitor; and so on.

\subsubsection{Time Costing for Training and Testing}

To train our DPL model, it takes $6$ hours in AlexNet and $28$ hours in VGG16.
During testing, our DPL only takes $1.85$s and $2.8$s per-image in AlexNet and VGG16 respectively.
It is much faster comparing with the HCP \cite{Ref:Wei2015}~($3$s and $10$s per-image in AlexNet and VGG16 respectively) that has achieved the state-of-the-art performance on object classification previously.

\begin{table}[!t]
\begin{center}
\caption{Object discovery results (CorLoc in $\%$) on PASCAL VOC 2012 trainval set.
$^{*}$ denotes results from unsupervised object co-localization methods.}
\label{table:dis_2012}
\resizebox{\linewidth}{!}{
\begin{tabular}{@{}L{3.2cm}|*{10}{x}|x}
\toprule
method & aero & bike & bird & boat & bottle & bus & car & cat & chair & cow &   \\
\midrule
Cho et al. \cite{Ref:Cho2015}$^{*}$ & $57.0$ & $41.2$ & $36.0$ & $26.9$ & $5.0$ & $81.1$ & $54.6$ & $50.9$ & $18.2$ & $54.0$ &\\
Li et al. \cite{Ref:Li2016eccv}$^{*}$ & $65.7$ & $57.8$ & $47.9$ & $28.9$ & $6.0$ & $74.9$ & $48.4$ & $48.4$ & $14.6$ & $54.4$ &\\
\midrule
DPL-AlexNet & $78.1$ & $59.0$ & $47.9$ & $22.9$ & $22.7$ & $76.1$ & $57.0$ & $39.0$ & $16.9$ & $57.9$ &\\
DPL-VGG16 & $79.0$ & $63.2$ & $50.3$ & $39.3$ & $38.3$ & $74.9$ & $58.1$ & $45.8$ & $21.9$ & $57.6$ &\\
\bottomrule
\toprule
method & table & dog & horse & mbike & persn & plant & sheep & sofa & train & tv & Ave.\\
\midrule
Cho et al. \cite{Ref:Cho2015}$^{*}$ & $31.2$ & $44.9$ & $64.8$ & $48.0$ & $13.0$ & $11.7$ & $51.4$ & $45.3$ & $64.6$ & $39.2$ & $41.8$\\
Li et al. \cite{Ref:Li2016eccv}$^{*}$ & $23.9$ & $50.2$ & $69.9$ & $68.4$ & $24.0$ & $14.2$ & $52.7$ & $30.9$ & $72.4$ & $21.6$ & $43.8$\\
\midrule
DPL-AlexNet & $23.1$ & $48.7$ & $59.2$ & $78.4$ & $36.0$ & $39.8$ & $65.6$ & $49.4$ & $49.5$ & $46.2$ & $48.7$\\
DPL-VGG16 & $38.9$ & $51.0$ & $43.9$ & $80.8$ & $16.5$ & $39.6$ & $58.3$ & $49.9$ & $57.3$ & $56.1$ & $51.0$\\
\bottomrule
\end{tabular}
}
\end{center}
\end{table}

\subsection{Object discovery}

We also perform some object discovery experiments. For object discovery, we only use the predicted patch scores from the discovery block in Figure~\ref{fig:DPL_architecture} and choose the patch with maximum score. The results on VOC 2007 and VOC 2012 are shown in Table~\ref{table:dis_2007} and Table~\ref{table:dis_2012}.

From the results, we can observe that our method can achieve quite competitive performance on object discovery. It outperforms other MIL-based methods like \cite{Ref:Wang2015,Ref:Cinbis2014}, but a little weaker than the method in \cite{Ref:Wang2014}. The method in \cite{Ref:Wang2014} finds a compact cluster for object and some clusters for the background. Except for being sensitive to the number of clusters, it is a must to tune parameters for each class tediously.
As other weakly supervised methods do not compute their CorLoc on VOC 2012, we only compare our method with unsupervised object co-localization methods \cite{Ref:Cho2015,Ref:Li2016eccv} in Table~\ref{table:dis_2012}.
We can observe that our method outperforms the co-localization methods \cite{Ref:Cho2015,Ref:Li2016eccv} on VOC 2012.
It is not surprise as our method benefits from image-level annotations, whereas \cite{Ref:Cho2015,Ref:Li2016eccv} are unsupervised (without image-level labels during training).

\begin{figure*}[!t]
\centering
\centerline{
\includegraphics[width=0.95\linewidth]{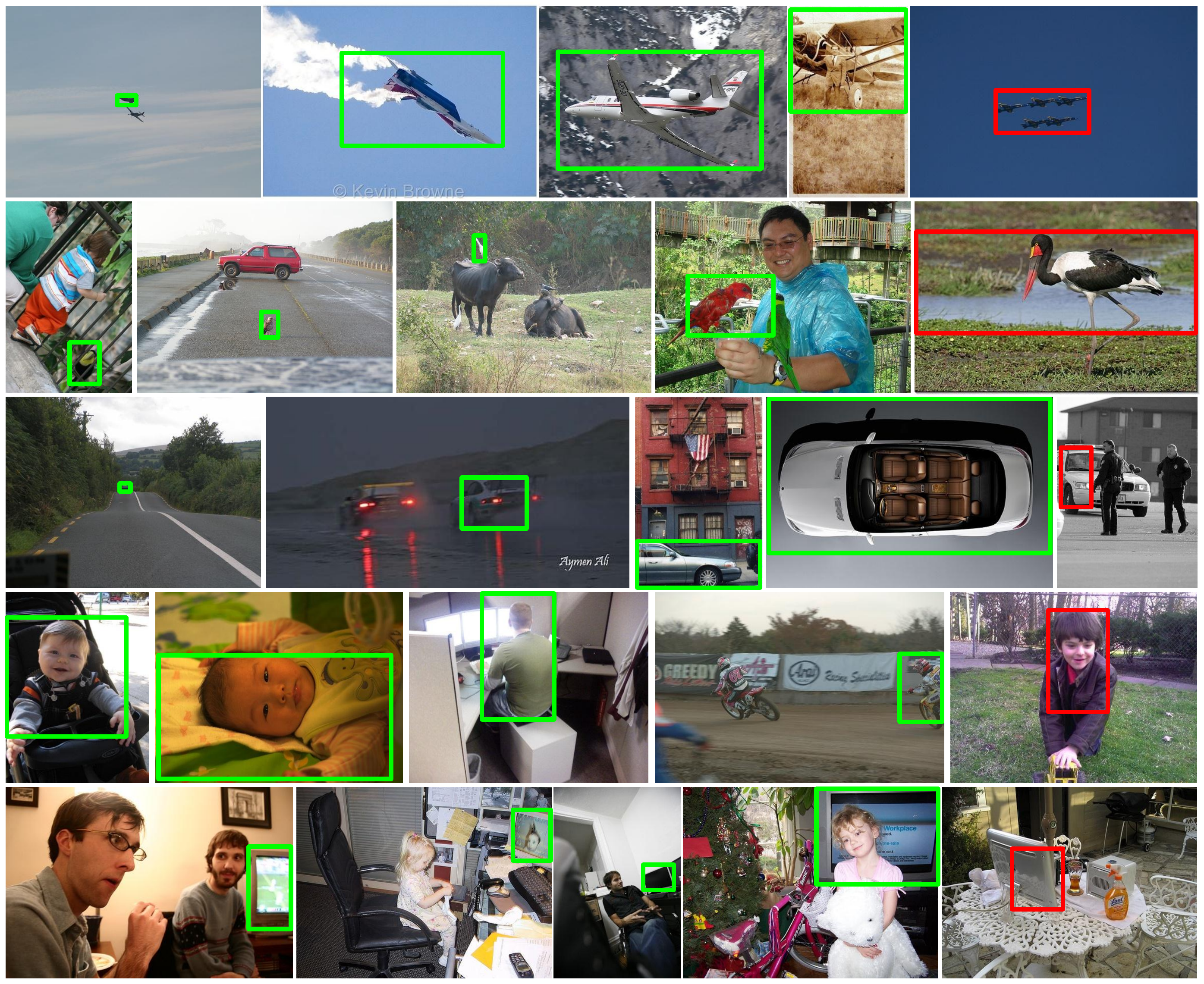}
}
\caption{Some discovery results on several classes on VOC 2007. Each row denotes one class. These classes are, from top row to bottom, aeroplane, bird, car, person, and tvmonitor. Green rectangle denotes success cases, and red rectangle denotes failure cases.}
\label{fig:dis_results}
\end{figure*}

Figure~\ref{fig:dis_results} shows some success and failure discovery cases on VOC 2007. As we can observe, though failure cases do not perform that well, they can still find the representative part of the whole object (e.g., the face for person), or the box not only including the object but also containing its adjacent similar objects and can locate the object exactly.

\section{Discussion}
\label{sec:discu}

In this section, we will discuss the influence factors of the multi-task learning, the image scales, and the method to generate patches. Without loss generality, we only choose AlexNet to perform experiments on the PASCAL VOC 2007 dataset. If not specified, all the reported testing time in this section does not includes that of the patch generation procedure.

\subsection{Multi-task vs. Single-Task}

Multi-task learning may improve the performance for each task as different tasks can influence each other by the shared representation \cite{Ref:Caruana1997}. Here we test the influence of different tasks. The results are shown in Table~\ref{table:task}. As we can see, the multi-task learning can improve the classification mAP by $0.3\%$ and the discovery CorLoc by $3.1\%$. These results demonstrate that the two tasks are complementary to some degree.

\begin{table}[!t]\footnotesize
\begin{center}
\caption{Results on PASCAL VOC 2007 for different tasks (mAP for Object classification and CorLoc for object discovery).}
\label{table:task}
\begin{tabular}{c | c c c}
\toprule
& multi-task & classification & discovery \\
\midrule
mAP ($\%$) & $\mathbf{85.3}$ & $85.0$ & $83.1$\\
CorLoc ($\%$) & $\mathbf{43.5}$ & - & $40.4$\\
testing time (s/image) & $0.35$ & $0.35$ & $0.35$\\
\bottomrule
\end{tabular}
\end{center}
\end{table}

\begin{table}[!t]\footnotesize
\begin{center}
\caption{Results on PASCAL VOC 2007 for different setups of image scale (mAP for Object classification and CorLoc for object discovery).}
\label{table:scale}
\begin{tabular}{c | c c}
\toprule
& multi-scale & single-scale \\
\midrule
mAP ($\%$) & $\mathbf{85.3}$ & $82.7$\\
CorLoc ($\%$) & $\mathbf{43.5}$ & $39.9$\\
testing time (s/image) & $0.35$ & $\mathbf{0.07}$\\
\bottomrule
\end{tabular}
\end{center}
\end{table}

\subsection{Multi-scale vs. Single-Scale}

To evaluate the influence of image scales, we conduct a single-scale experiment that only uses one scale $600$ to compare with the five scales experiment. The results are shown in Table~\ref{table:scale}. We can observe that, multi-scale can improve the classification and discovery results evidently ($+2.6$ and $+3.6$ respectively) but with the additional testing time.
Notably, using a multi-scale approach allows one to increase the accuracy for both tasks.
Even though this approach increases the testing time slightly, it could be of interest for applications in which accuracy remains more important than response time during system operation.

\begin{table}[!t]\footnotesize
\begin{center}
\caption{Results on PASCAL VOC 2007 for different patch generation methods and HCP \cite{Ref:Wei2015} (mAP for Object classification and CorLoc for object discovery). $^{+}$ denotes the time costing with the addition of patch generation procedure.}
\label{table:patch}
\begin{tabular}{c | c c c | c}
\toprule
\multirow{2}{1cm}{} & \multicolumn{3}{c|}{DPL} & \multirow{2}{2cm}{HCP \cite{Ref:Wei2015}}\\ & SS & EdgeBoxes & SW &\\
\midrule
mAP ($\%$) & $\mathbf{85.3}$& $85.2$ & $\mathbf{85.3}$ & $82.7$\\
CorLoc ($\%$) & $\mathbf{43.5}$ & $40.0$ & $24.4$ & -\\
testing time (s/image) & $0.35$ & $\mathbf{0.15}$ & $0.22$ & $2.75$\\
testing time$^{+}$ (s/image) & $1.85$ & $0.4$ & $\mathbf{0.22}$ & $3$\\
\bottomrule
\end{tabular}
\end{center}
\end{table}

\subsection{The Influence of Different Patch Generation Methods}

In the previous experiments, we choose SS \cite{Ref:Uijlings2013} to extract patches. Here we will compare three different methods to generate patches, including SS \cite{Ref:Uijlings2013}, EdgeBoxes \cite{Ref:Zitnick2014}, and Sliding Window~(SW). For EdgeBoxes, we generate $256$ patches for each image, so it can accelerate the testing speed (we also test the performance when increase the patch number, but the results show that the performance is only improved a little but the speed slow down a lot). For SW method, we extract patches from $7$ different scales widow $32 \times \{2, 3, ..., 8\}$ with step size $32$. This operation will generate $500$ to $1000$ patches per-image. The results are shown in Table~\ref{table:patch}. From the results, we can observe that the method to extract patches affects the performance greatly, especially for object discovery. What is more, SS is the best method for both object classification and discovery. It is interesting that the SW method can achieve similar classification mAP comparing with SS with less testing time. Notice that the time to generate SW patches is negligible, so during testing, the SW method is about $2\times$, $8\times$, and $13\times$ faster than EdgeBoxes, SS, and HCP, respectively.
For systems only focusing on object classification, the SW method is preferable as it reduces the testing time significantly with no cost of performance.

\section{Conclusions}
\label{sec:conclu}

In this paper, a novel DPL method is proposed, which integrates the patch feature learning, image representation learning, object classification and discovery into a unified framework. 
The DPL explicitly optimizes patch-level image representation, which is totally different from conventional CNNs.
It also combines the CNN based patch-level feature learning with MIL methods, thus can be trained in a weakly supervised manner.
The excellent performance of DPL on object classification and discovery confirms its effectiveness. 
These inspiring results show that learning good patch-level image presentation and combining CNNs with MIL are very promising directions to explore in various vision problems. 
In the future, we will study how to apply DPL for other visual recognition problems, including introducing DPL into solve very large scale problems.

\section*{Acknowledgements}

This work was primarily supported by National Natural Science Foundation of China (NSFC) (No. 61503145, No. 61572207, and No. 61573160) and the CAST Young Talent Supporting Program.

\section*{References}
{\scriptsize
\bibliographystyle{elsarticle-num}
\bibliography{egbib}
}

\end{document}